%% file: main.tex
\documentclass[10pt,twocolumn,letterpaper]{article}

\usepackage{cvpr}              

\input{preamble}

\definecolor{cvprblue}{rgb}{0.21,0.49,0.74}
\usepackage[pagebackref,breaklinks,colorlinks,allcolors=cvprblue]{hyperref}

\def\paperID{*****} 
\def\confName{CVPR}
\def\confYear{2026}

\title{3rd Place of MeViS-Audio Track of the 5th PVUW: VIRST-Audio}

\author{
Jihwan Hong$^{1}$ \quad Jaeyoung Do$^{1,2,}$\thanks{Corresponding author}\\
AIDAS Laboratory, $^1$IPAI \& $^2$ECE, Seoul National University \\
{\tt\small \{csjihwanh, jaeyoung.do\}@snu.ac.kr}\\
Team: SNU\_AIDAS
}

\begin{document}
\maketitle
\input{sec/0_abstract}    
\input{sec/1_intro}
\input{sec/2_related}
\input{sec/3_method}
\input{sec/4_experiments}
\input{sec/5_conclusion}
{
    \small
    \bibliographystyle{ieeenat_fullname}
    \bibliography{main}
}


\end{document}

%% file: preamble.tex
\newcommand{\methodname}{VIRST-Audio}



%% file: sec/0_abstract.tex
\begin{abstract}
Audio-based Referring Video Object Segmentation (ARVOS) requires grounding audio queries into pixel-level object masks over time, posing challenges in bridging acoustic signals with spatio-temporal visual representations. In this report, we present \textbf{\methodname{}}, a practical framework built upon a pretrained RVOS model integrated with a vision-language architecture. Instead of relying on audio-specific training, we convert input audio into text using an ASR module and perform segmentation using text-based supervision, enabling effective transfer from text-based reasoning to audio-driven scenarios. To improve robustness, we further incorporate an \textbf{existence-aware gating} mechanism that estimates whether the referred target object is present in the video and suppresses predictions when it is absent, reducing hallucinated masks and stabilizing segmentation behavior. We evaluate our approach on the MeViS-Audio track of the 5th PVUW Challenge, where \textbf{\methodname{}} achieves 3rd place, demonstrating strong generalization and reliable performance in audio-based referring video segmentation. Code is available at \url{https://github.com/AIDASLab/VIRST/tree/virst-audio}.
\end{abstract}

%% file: sec/1_intro.tex
\section{Introduction}

Video Object Segmentation (VOS)~\cite{caelles2017osvos,MOSEv2,perazzi2016davis-vos} aims to understand videos at the pixel level by identifying and segmenting target objects over time. Building upon this, recent works have extended VOS toward more realistic and flexible settings by incorporating diverse input modalities, including text~\cite{khoreva2018ref-davis17,ding2023mevis,gavrilyuk2018actor,MeViSv2}, user interactions~\cite{ravi2024sam2,carion2025sam3}, and audio~\cite{zhou2022avsbench,liu2024luavs,MeViSv2}. Such multimodal extensions require models to bridge high-level semantic inputs with fine-grained spatio-temporal representations, making pixel-level video understanding significantly more challenging. Addressing this challenge is crucial for real-world applications, where models must reliably interpret diverse inputs and produce precise, temporally consistent segmentation under unconstrained conditions.

To address these challenges, the Pixel-level Video Understanding in the Wild (PVUW) workshop is held annually to promote research on realistic video-centric segmentation. The associated challenge consists of three tracks: (1) Complex Video Object Segmentation (MOSEv2~\cite{MOSEv2}), which focuses on object tracking and segmentation in complex environments; (2) Text-based Referring Motion Expression Video Segmentation (MeViSv2-Text~\cite{MeViSv2}), which targets segmenting objects based on motion descriptions in natural language; and (3) Audio-based Referring Motion Expression Video Segmentation (MeViSv2-Audio~\cite{MeViSv2}), which extends this setting to audio queries, requiring models to ground motion descriptions from acoustic signals.

We focus on Audio-based Referring Video Object Segmentation (ARVOS) in this paper. While general audio-guided segmentation may involve diverse acoustic signals, the ARVOS setting is largely speech-driven, where the input audio conveys semantic descriptions of target objects. Directly adopting audio-encoding models is a straightforward approach, but it often limits adaptability across modalities. 

To address this, we introduce \textbf{\methodname{}}, a framework that leverages a Referring Video Object Segmentation (RVOS) expert model integrated with a vision-language model, without requiring any ARVOS-specific training data. By converting speech into text via an ASR module, the task is reformulated as text-based referring segmentation. As a result, \methodname{}, trained solely on text-based RVOS data, generalizes effectively to the audio-driven setting without additional supervision. This design achieves 3rd place on the Audio-based Referring Motion Expression Video Segmentation track.

Plus, to mitigate hallucination in VOS, N-acc. (No-target accuracy) and T-acc. (Target accuracy)~\cite{liu2023gres} are introduced in MeViS-Audio~\cite{MeViSv2}. Specifically, N-acc. is defined as $\frac{TN}{TN+FP}$, measuring how accurately the model predicts the absence of the target, while T-acc. is defined as $\frac{TP}{TP+FN}$, reflecting how well the model identifies the presence of the target object. 

Motivated by this, we introduce an \textbf{existence-aware gating} mechanism that explicitly models the presence of the target object in the video, enabling more robust segmentation while mitigating hallucinated predictions. During inference, confidence-based thresholding is applied to suppress segmentation when the target is predicted to be absent. This simple yet effective design reduces false positives and leads to consistent performance improvements.

%% file: sec/2_related.tex
\section{Related Works}
\subsection{Referring Video Object Segmentation}

Referring Video Object Segmentation (RVOS) aims to segment target objects in videos conditioned on natural language descriptions~\cite{gavrilyuk2018actor,khoreva2018ref-davis17}. Unlike conventional VOS, RVOS requires aligning linguistic queries with spatio-temporal visual content, making both semantic understanding and temporal consistency essential. The task has been extensively studied on benchmarks such as Ref-YouTube-VOS~\cite{seo2020urvos}, MeViS~\cite{ding2023mevis}, ReVOS~\cite{yan2024visa}, and MeViSv2~\cite{MeViSv2}, which introduce increasing levels of complexity in terms of motion expressions, compositional queries, and real-world variability.

Early approaches~\cite{botach2022mttr,seo2020urvos,ding2023mevis,wu2022referformer} typically employed separate visual and language encoders followed by lightweight mask decoders, limiting their ability to capture complex semantic relationships. More recent methods leverage vision-language models (VLMs) to improve cross-modal reasoning and grounding. For instance, VISA~\cite{yan2024visa} integrates VLM representations with SAM~\cite{kirillov2023sam1} for keyframe segmentation and propagates masks using XMem~\cite{cheng2022xmem}. In parallel, VIRST~\cite{hong2026virst} introduces a unified framework that combines global semantic reasoning with pixel-level segmentation through spatio-temporal fusion, enabling more robust performance under complex and ambiguous queries. These advances reflect a broader trend toward tightly coupled multimodal architectures for scalable and reliable RVOS.

\subsection{Audio-guided Video Object Segmentation}

Audio-guided Video Object Segmentation (AVOS) aims to segment objects in videos based on audio signals that are temporally aligned with visual content. Unlike RVOS, which relies on explicit linguistic queries, AVOS typically uses general acoustic cues such as object sounds or environmental noise, making the task inherently ambiguous. Benchmarks such as AVSBench~\cite{zhou2022avsbench} and LU-AVS~\cite{liu2024luavs} highlight challenges including noisy audio, multiple sound sources, and weak correspondence between audio and visual signals.

Recent approaches learn audio-visual alignment through joint representations and attention mechanisms~\cite{zhou2022avsbench,liu2024luavs}, and more recent works adopt transformer-based architectures for stronger temporal reasoning~\cite{huang2025vct}. AVOS is fundamentally different from speech-based settings, where audio provides explicit semantic descriptions. In such cases, speech can be converted into text and addressed as a referring segmentation problem, enabling the use of text-based reasoning models as in ARVOS.

%% file: sec/3_method.tex
\section{Method}
\subsection{Problem Formulation}

In the ARVOS task, given a video $V \in \mathbb{R}^{H \times W \times C \times T}$ and an audio query $a$ describing a set of target objects $\mathcal{O} = \{o_i\}_{i=1}^N$ in the video, we aim to predict the binary mask of the object set. Each object $o_i \in \mathcal{O}$ is associated with a binary segmentation mask $\mathcal{M}_{o_i} \in \mathbb{R}^{H \times W \times T}$. The target mask is defined as the union of all object masks:
\begin{equation}
    \mathcal{M}_{\mathcal{O}} = \bigcup_{o_i \in \mathcal{O}} \mathcal{M}_{o_i}.
\end{equation}

\subsection{\methodname{}}

\input{fig/tex/overall_arch}

We propose \textbf{\methodname{}}, which builds upon VIRST (Video-Instructed Reasoning Assistant for Spatio-Temporal Segmentation)~\cite{hong2026virst}. The overall pipeline is illustrated in Fig.~\ref{fig:overall_arch}. 

VIRST is a VLM-based segmentation framework that combines global semantic understanding with pixel-level segmentation. It integrates CLIP~\cite{cherti2023openclip}-based video encoder features and SAM2~\cite{ravi2024sam2} encoder features via an ST-Fusion module~\cite{hong2026virst}, which performs cross-attention using a learnable token \texttt{[ST]}. The fused representation is then used as a prompt for the SAM2 mask decoder.

To extend this framework to ARVOS, \methodname{} incorporates an ASR (Automatic Speech Recognition) module that converts the input audio into text. The transcribed text is then used as the language input to the VLM. Notably, \methodname{} achieves strong performance on ARVOS benchmarks \emph{without any fine-tuning on ARVOS datasets}, demonstrating effective transfer of text-based reasoning to the audio domain.

\subsection{Existence-Aware Segmentation Gating}

Robust segmentation in the absence of the target object is a critical challenge in RVOS, as false positives can significantly limit real-world applicability and incur unnecessary computational cost. Recent works~\cite{yan2024visa, li2023r2vos,MeViSv2} have increasingly focused on mitigating false positive and false negative predictions. To address this issue, we introduce an \textbf{existence-aware gating mechanism} that determines whether the audio-referred target object exists in the video and suppresses spurious segmentation when the target is absent.

For this, we define an indicator function that determines whether the referred target object is present in the video:
\begin{equation}
\mathbb{I}(V, a) =
\begin{cases}
1, & \text{if } \exists\, t \text{ such that } \mathcal{M}_{\mathcal{O}}(:,:,t) \neq \mathbf{0}, \\
0, & \text{otherwise},
\end{cases}
\end{equation}
where $\mathcal{M}_{\mathcal{O}}$ denotes the union mask of the audio-referred target objects in video $V$.

From the ST-Fusion model output $\mathbf{F} \in \mathbb{R}^{N \times T \times D}$, which is used as prompts for the mask decoder, we apply a lightweight existence prediction module:
\begin{equation}
z = f_{\text{exist}}(\mathbf{F}), \quad p = \sigma(z),
\end{equation}
where $p$ denotes the probability that the referred target exists in the video.

For training, we supervise the existence prediction using a binary cross-entropy (BCE) loss, where the target label is defined based on whether the ground-truth mask is non-empty. In particular, samples without the referred object (i.e., $\mathbb{I}(V,a)=0$) are excluded from segmentation supervision to avoid learning spurious mask predictions.

At inference time, the predicted existence probability $p$ is compared with a threshold $\tau$ to determine whether segmentation should be performed. If $p < \tau$, the model directly outputs an empty prediction without invoking the segmentation module. Otherwise, segmentation is conducted using the predicted prompts and propagated across the video.

%% file: fig/tex/overall_arch.tex
\begin{figure*}[t]
  \centering
    \includegraphics[width=0.9\linewidth]{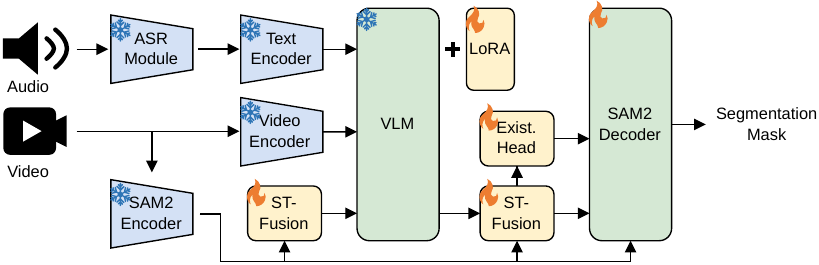} 
    \caption{
    \textbf{Overall architecture of \methodname{}.}
    }

    \label{fig:overall_arch}
\end{figure*}

%% file: sec/4_experiments.tex
\section{Experiments}
\subsection{Implementation Details}

The overall architecture is based on VIRST~\cite{hong2026virst}, where we adopt Whisper-Large~\cite{radford2023whisper} as the ASR module. We initialize the model with pretrained VIRST weights trained on multiple datasets, including ReVOS~\cite{yan2024visa}, MeViSv1~\cite{ding2023mevis}, Ref-YouTube-VOS~\cite{seo2020urvos}, and Ref-DAVIS17~\cite{khoreva2018ref-davis17}. 

We then fine-tune the model on the MeViSv2-Text~\cite{MeViSv2} split, updating only the ST-Fusion module, LoRA layers, and SAM2 memory and decoder modules, while keeping all other components frozen. Training is conducted on 4 A100 GPUs, and inference is performed on a single A100 GPU. 

\subsection{Experiemental Results}
\subsubsection{MeViS-Audio Track of the 5th PVUW Results}

Table~\ref{tab:mevis_audio_results} presents the results on the MeViS-Audio track of the 5th PVUW Challenge. Our method ranks 3rd out of 13 participating teams, achieving a $\mathcal{J\&F}$ score of 0.54, with $\mathcal{J}$ and $\mathcal{F}$ scores of 0.52 and 0.56, respectively. These results demonstrate that our approach produces consistent segmentation quality in terms of both region similarity and boundary accuracy.

Qualitative results in Fig.~\ref{fig:qualitative} further support these observations. In the multi-object scenario (a), \methodname{} successfully distinguishes and segments the correct target among multiple candidates. When the target object is clearly specified, as in (c) and (d), the model produces accurate and consistent segmentation results across frames. Importantly, in the case where the referred object does not exist, as shown in (b), the model correctly outputs no segmentation, effectively suppressing false positive predictions. This behavior highlights the effectiveness of the proposed existence-aware gating in handling both ambiguous and negative queries.

Beyond mask quality, our method achieves balanced performance in both N-acc. and T-acc., indicating reliable behavior across both negative and positive cases. In particular, the model is able to effectively suppress predictions when the referred target is absent, while maintaining strong segmentation performance when the target is present. This suggests that the proposed existence-aware segmentation gating provides a useful global prior for filtering invalid queries and stabilizing the overall prediction pipeline.

Notably, although the model is trained using only text-based supervision, it generalizes well to audio-driven queries via speech-based descriptions. This indicates effective cross-modal knowledge transfer from text to audio, enabled by the integration of the ASR module. The results suggest that the learned representation can bridge linguistic modalities while maintaining robust segmentation performance.

\input{tab/challenge_result}

\subsubsection{Existence-Aware Gating Ablation}
\input{tab/gating_ablation}

Table~\ref{tab:gating_ablation} presents an ablation study on the proposed existence-aware segmentation gating with different thresholds $\tau$. We report $\mathcal{J}$, $\mathcal{F}$, $\mathcal{J\&F}$, and the hallucination-aware metrics N-acc. and T-acc., where N-acc. measures the ability to correctly identify the presence of the target object, and T-acc. evaluates the ability to correctly suppress predictions when the object is absent.

Compared to the VIRST baseline, introducing existence-aware gating consistently improves segmentation quality, as reflected by the increase in $\mathcal{J\&F}$ from 0.49 to 0.54. At the same time, gating significantly improves N-acc., indicating better handling of positive samples. We also observe a trade-off between N-acc. and T-acc. as the threshold $\tau$ increases. Specifically, a higher threshold (e.g., $\tau=0.9$) leads to stronger suppression of false positives, resulting in higher N-acc., while slightly reducing T-acc., which reflects a more conservative prediction behavior.

Overall, $\tau=0.8$ provides the best balance between segmentation quality and hallucination robustness, achieving the highest Final score. These results demonstrate that the proposed gating mechanism effectively controls spurious predictions while maintaining strong segmentation performance.

\input{fig/tex/qualitative}

%% file: tab/challenge_result.tex
\begin{table}[t]
\centering
\setlength{\tabcolsep}{4pt}
\renewcommand{\arraystretch}{0.95}
\footnotesize
\caption{Results on the PVUW 2026 MeViS-Audio track. \methodname{} ranks 3rd among 13 teams. Our results are \textbf{bolded}.}
\begin{tabular}{lccccccc}
\toprule
\textbf{Rank} & \textbf{Method} & $\mathcal{J\&F}$ & $\mathcal{J}$ & $\mathcal{F}$ & N-acc. & T-acc. & Final \\
\midrule
1 & yahooo & 0.67 & 0.64 & 0.70 & 0.89 & 0.98 & 0.85 \\
2 & wangzhiyu918 & 0.64 & 0.61 & 0.67 & 0.83 & 0.95 & 0.81 \\
3 & \textbf{Ours} & \textbf{0.54} & \textbf{0.52} & \textbf{0.56} & \textbf{0.70} & \textbf{0.82} & \textbf{0.68} \\
4 & vvv666 & 0.47 & 0.44 & 0.50 & 0.12 & 0.98 & 0.52 \\
5 & liyiying & 0.48 & 0.45 & 0.50 & 0.09 & 0.97 & 0.51 \\
6 & eronguyen & 0.24 & 0.23 & 0.26 & 0.18 & 0.91 & 0.45 \\
7 & WDL & 0.30 & 0.26 & 0.34 & 0.00 & 1.00 & 0.43 \\
\bottomrule
\end{tabular}
\label{tab:mevis_audio_results}
\end{table}

%% file: tab/gating_ablation.tex
\begin{table}[t]
\centering
\setlength{\tabcolsep}{5pt}
\renewcommand{\arraystretch}{0.95}
\footnotesize
\caption{Effect of existence-aware gating with different thresholds $\tau$. Final score is the average of $\mathcal{J\&F}$, N-acc., and T-acc. Best results are \textbf{in bold}.}
\begin{tabular}{lcccccc}
\toprule
\textbf{Method} & $\mathcal{J\&F}$ & $\mathcal{J}$ & $\mathcal{F}$ & N-acc. & T-acc. & \textbf{Final} \\
\midrule
VIRST (baseline) & 0.49 & 0.47 & 0.51 & 0.62 & 0.84 & 0.65 \\
w/o gating       & 0.52 & 0.49 & 0.54 & 0.58 & \textbf{0.89} & 0.66 \\
gating ($\tau=0.8$) & \textbf{0.54} & \textbf{0.52} & \textbf{0.56} & 0.70 & 0.82 & \textbf{0.68} \\
gating ($\tau=0.9$) & 0.53 & 0.51 & 0.55 & \textbf{0.71} & 0.78 & 0.67 \\
\bottomrule
\end{tabular}
\label{tab:gating_ablation}
\end{table}

%% file: fig/tex/qualitative.tex
\begin{figure}[t]
  \centering
    \includegraphics[width=\linewidth]{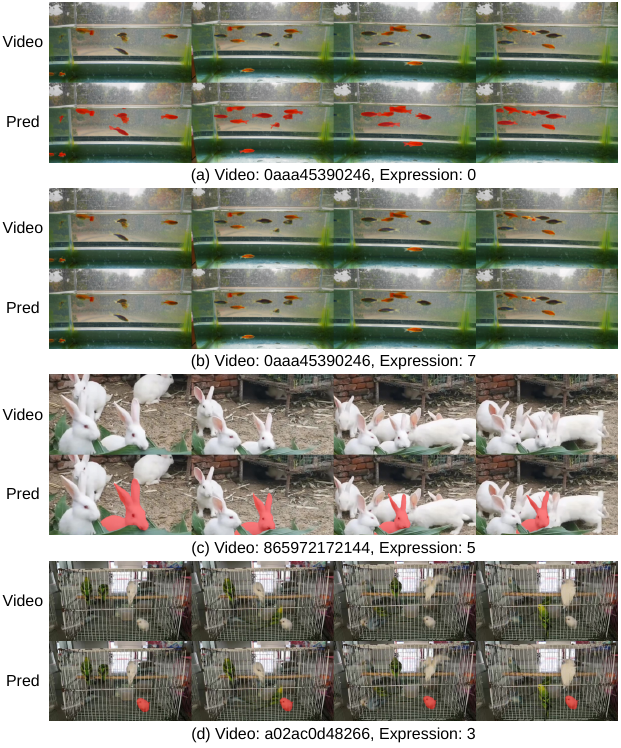} 
    \caption{
    Qualitative results of \methodname{} on the MeViS-Audio test set.
    }
    \label{fig:qualitative}
\end{figure}

%% file: sec/5_conclusion.tex
\section{Conclusion}

In this report, we presented \methodname{}, a practical framework for Audio-based Referring Video Object Segmentation built upon a pretrained RVOS model and a vision-language architecture. By converting speech into text via an ASR module, our approach reformulates ARVOS as a text-based referring segmentation problem, enabling effective transfer from text-based supervision to audio-driven scenarios without requiring any audio-specific training. This design leverages existing multimodal reasoning capabilities while maintaining a simple and scalable pipeline.

To improve robustness, we introduced an existence-aware gating mechanism that explicitly models whether the target object is present in the video. This mechanism suppresses predictions when the target is absent, reducing hallucinated masks and improving reliability under challenging conditions. Through quantitative and qualitative evaluations on the MeViS-Audio track of the 5th PVUW Challenge, \methodname{} demonstrates consistent segmentation quality and balanced performance across both positive and negative cases. Overall, the results highlight the effectiveness of combining cross-modal transfer with explicit existence modeling for robust audio-based video object segmentation.